\documentclass[letterpaper]{article} 
\usepackage{aaai2026}  
\usepackage{times}  
\usepackage{helvet}  
\usepackage{courier}  
\usepackage[hyphens]{url}  
\usepackage{graphicx} 
\usepackage{natbib}  
\usepackage{caption} 
\usepackage{algorithm}
\usepackage{algorithmic}
\usepackage{float}
\usepackage{tabularray}
\usepackage{subcaption}
\usepackage{amsmath}

\usepackage{newfloat}
\usepackage{listings}
\DeclareCaptionStyle{ruled}{labelfont=normalfont,labelsep=colon,strut=off} 
\floatstyle{ruled}
\newfloat{listing}{tb}{lst}{}
\floatname{listing}{Listing}
\title{Text-to-Vector Conversion for Residential Plan Design}
\author {
    Egor Bazhenov\textsuperscript{\rm 1},
    Stepan Kasai\textsuperscript{\rm 1},
    Viacheslav Shalamov\textsuperscript{\rm 1},
    Valeria Efimova\textsuperscript{\rm 1}
}
\affiliations {
    \textsuperscript{\rm 1}ITMO University, Saint-Petersburg, Russia\\
    egorbazhenov@niuitmo.ru, kasai.stepa@gmail.com, shalamov@itmo.ru, vefimova@itmo.ru
}

\begin{document}

\maketitle

\begin{abstract}
Computer graphics, comprising both raster and vector components, is a fundamental part of modern science, industry, and digital communication. 
While raster graphics offer ease of use, its pixel-based structure limits scalability. 
Vector graphics, defined by mathematical primitives, provides scalability without quality loss, however, it is more complex to produce. 
For design and architecture, the versatility of vector graphics is paramount, despite its computational demands. 
This paper introduces a novel method for generating vector residential plans from textual descriptions. 
Our approach surpasses existing solutions by approximately $5\%$ in CLIPScore-based visual quality, benefiting from its inherent handling of right angles and flexible settings.
Additionally, we present a new algorithm for vectorizing raster plans into structured vector images.
Such images have a better CLIPscore compared to others by about $4\%$.
\end{abstract}


\section{Introduction}

Vector and raster graphics belong to the domain of computer graphics that plays an important role in manufacturing, marketing, design, architecture and digital communications. 
Raster images consist of pixels and have many formats, the most common being PNG, BMP, JPG, and JPEG. 
This type of graphics is easy to create and edit; moreover, it is supported by numerous software products. 
However, raster graphics lacks scalability, and the image weight depends quadratically on the image quality. 
Vector graphics consists of paths that represent various geometric shapes such as squares, circles, lines, arcs, and others. 
Such images are scalable and have small file size. 
The main disadvantage of vector graphics is that such images are difficult to create and edit, as they require an understanding of the structure of SVG~\cite{quint2003scalable}. 
Nevertheless, designers and architects prefer this type of graphics for developing residential plans.

Global urbanization of population~\cite{ichimura2003urbanization} makes residential planning critical for real estate development.
Housing construction is time consuming  and investment demanding with residential plans design being only a part of it.
Thus, the task of automating the residential plans generation is extremely relevant against the background of the growing demand to save developers' time and resources.
Currently, there are two main approaches to generate vector residential plans based on textual description. 
The first one comprises a generator and a vectorizer. 
The main idea is that a generator creates a raster residential plan based on a textual description, and then a vectorizer converts a raster image to a vector one.
The second approach is to generate a vector plan using a Large Language Model (LLM)~\cite{raiaan2024review}. 
Since SVG images consist of code, this approach can also be used to generate residential plans. 

In our paper, a raster-generation-vectorization approach is chosen as the baseline.
Applying this approach, we can customize and configure the generative algorithm.
We have chosen Stable Diffusion XL~\cite{podell2023sdxl} with proposed white loss as the main model for raster generation. 
For floor plan vectorization, we have developed the new algorithm based on angle detection and analytical analysis methods.

As a result, we have developed the algorithm for generating vector residential plans based on textual description. 
Specifically, the main contributions are as follows:
\begin{enumerate} 
  \item We propose the novel method for generating vector residential plans based on textual description that provides residential plans in SVG file format without extra paths.
  \item To facilitate further vectorization, we suggest the method to adapt a pre-trained image generation model to the generation of raster images with a similar structure to vector ones.
  \item We have developed the new algorithm for plan vectorization that allows obtaining right angles and straight walls in a resulting vector representation, creating a structured vector image based on postprocessing analysis.
\end{enumerate}

We publicly release our code\footnote{\url{https://github.com/CTLab-ITMO/GenPlan}} to support further research and comparative evaluation.

\section{Related works}\label{sec:releated}

The problem of generating a vector image based on the textual description can be solved by combining a raster generator with a vectorizer or end-to-end vector generator. 
With benefits and drawbacks these approaches have, the current open source algorithms do not seem to be effective or efficient in residential plan generation. 

\subsection{Raster Generation with Vectorization}

The raster generator creates a bitmap plan based on a textual description, and then the vectorization algorithm converts the bitmap image to a vector format.
Due to the flexibility of this approach, we can configure and select the generator and vectorizer best suited for the task.
This section describes various algorithms for raster generation and vectorization, as well as their features for the vector plans generation task.

\subsubsection{Raster Generation}

For the task of raster generation based on the textual description the two main approaches are often applied: Rule-based generation models and Neural networks. 
Each approach has its advantages and disadvantages in terms of the given raster plan generation task.

\textit{Rule-based generation} for the task of generating raster images is deterministic and simple~\cite{wang2023rule,avila2021house} approach.
An image is selected based on the single textual template that can be customized depending on the task.
As all possible variations of the images are created in advance, this approach works quickly and transparently. 

Rule-based generation models have significant disadvantages. 
First, they lack diversity due to a limited number of ready-made raster images. 
Second, they lack scalability and flexibility, as completely new images cannot be created.
These disadvantages make Rule-based generation models narrowly applicable. 

\textit{Stable Diffusion}~\cite{rombach2022high} is one of the prominent models for text-to-image generation based on neural networks.
This model consists of three main parts: the text encoder (CLIPText model~\cite{qin2023cliptext}), the image information creator based on the U-Net model, and the image decoder of an autoencoder trained on image datasets.

Stable Diffusion is one of the most powerful models for image generation, but for the task of generation raster plans based on textual description this model application has some problems.
One of them is poor quality of generated plans due to this model lacks specialization.
This model also has problems with generation the right angles.

\textit{Stable Diffusion XL (SDXL)}~\cite{podell2023sdxl} is an upgraded version of Stable Diffusion with a larger model size, better prompt understanding. 
A larger U-Net backbone ($2.6$ billion parameters vs. SD 1.5’s $860$ Million) and a dual-text-encoder system (CLIP ViT-L + OpenCLIP ViT-bigG) allows significant improving the quality of generated plans.

SDXL generates images with high quality, which makes it suitable for plan generation. 
However, like Stable Diffusion, this model cannot overcome the problem with the right angles. 
Despite this, we chose this model for our approach as one of raster image generators due to its excellent quality of images.

\subsubsection{Vectorization}

Image vectorization has a lot of solutions, but every algorithm has specific characteristic.
This section presents the most popular and powerful vectorization algorithms, as well as their main features.

\textit{Vectorization algorithms based on machine learning}.
DiffVG~\cite{li2020differentiable} is an advanced vectorization algorithm that uses differentiable rasterization to gradually optimize a vector image to match a initial bitmap. 
It takes the input raster image and a number of paths in the final vector image, then iteratively refines the vector parameters via backpropagation based on a loss function. 
A different approach is taken by LIVE~\cite{ma2022towards}, which iteratively adds paths to the vector image. 
It identifies the largest monochrome area in the bitmap, adds a new path, and optimizes it using dual loss functions for similarity and to avoid self-intersection.

DiffVG and LIVE show low productivity and manages the task of vectorizing raster plan only partially.
One of the problems is an initial parameter --- the number of paths in the final vector image, since this value may be unknown.
Besides, these algorithms produce odd angles instead of the right ones resulting in the final image that only partially corresponds to the initial raster image. 
The problems above make DiffVG and LIVE unsuitable in plan vectorization.

\textit{Deterministic vectorization algorithms}.
There exist many deterministic vectorization algorithms based on tracing~\cite{ali2023survey}.
They work quickly due to the static analysis of the original bitmap image.
The main disadvantage of this approach results from the creation of redundant paths in the vector image; the structure of such an image becomes overloaded and hardly manually editable.

For the task of vectorizing plans, this approach creates an unstructured vector image that is difficult to process.
Despite fast operation, these algorithms are inconvenient to use for the task.

\subsection{Large Language Models}

Another approach for generating vector images is Large Language Models (LLM)~\cite{raiaan2024review}.
LLMs process the text and create a new one based on it. 
Since vector images consist of SVG code, which is XML markup, LLMs generate a vector image in one iteration, therefore, this process is extremely fast. 
Unfortunately, the key disadvantage of existing LLMs is the low quality of generated vector images.
SVG images have a complex structure, so current LLMs cannot efficiently manage the vector image generation task.
Numerous papers report on the poor quality of vector image generation using LLMs~\cite{cai2023leveraging}.

In our research we use LLMs for comparison.
They work extremely fast but generate plans with a primitive structure. 
The approach is promising for task of vector plans generation based on textual description, however, current LLMs cannot fully tackle the challenge.

\section{Methods}\label{sec:methods}

\begin{figure}[ht]  
    \centering
    \includegraphics[width=1.0\linewidth]{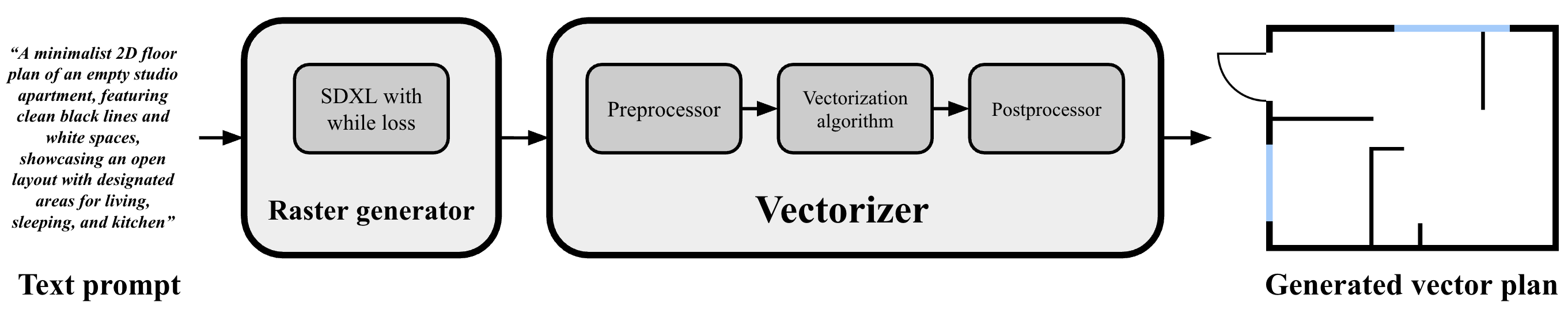}  
    \caption{Example of vector plan generation by our algorithm.}  
    \label{fig:pipline}  
\end{figure}

Our paper presents an approach consisting of a raster generator and a vectorizer.
SDXL with white loss was selected as model for raster generation.
For vectorization we have developed a new approach based on the Shi-Tomasi corner detection method~\cite{shi1994good} that can work with the right angles and straight walls.
Figure~\ref{fig:pipline} demonstrates the basic working pipeline of our algorithm.

\subsection{Generation}\label{sec:generator}

The first part of our approach is a raster generator. 
Among the model's limitations, it is the white background of bitmap images that is also a challenge: the models have difficulties with generating images on a white background. 
We assume that the issue results from the ambiguity of a prompt for image generation. 
More detailed prompts can contribute to handling the issue although their development is time-consuming. 
However, even a highly detailed prompt does not guarantee that white background will be generated. 
The image background may be in shades of gray, but for the human eye the difference is significant. 

We designed new loss function~\ref{eq:white_loss} to generate images on a white background:
\begin{equation}
    L_{white} = s \left\| D(p(x_0 | x_t)) - \mathrm{mask}(D(p(x_0 | x_t))) \right\|^2,
    \label{eq:white_loss}
\end{equation}
where $D$ --- SDXL decoder, $s$ --- scale, $x_0$ --- vector without noise in latent space, $x_t$ --- vector in latent space with noise corresponding to step $t$, $mask$ --- mask that defines the field that we would like to be white. 

We also need to update $x_t$ after this by following recurrent formula~\ref{eq:latent_update}:
\begin{equation}
    x_t = x_t - \mathrm{mask}_{\text{latent}} \left( \Delta x_t \cdot
    \left(1-\overline{\alpha_t}\right)
    /\ \overline{\alpha_t} \right),
    \label{eq:latent_update}
\end{equation}
where $t$ --- diffusion step, $\Delta x_t$ --- gradient of $x_t$ according to  $L_{\text{white}}$, $mask_{\text{latent}}$ ---
mask that is scaled to be used in latent space, $\overline{\alpha_t}$ --- cumulative product~\cite{rombach2022high}.

We use the SDXL decoder to obtain an image from $x_t$ on early time step $t$, then we apply a mask to this image and calculate the MSE loss, which is high in the mask field. 
Then, we update $x_t$ according to formula~\ref{eq:latent_update}.

\subsection{Vectorization}\label{sec:vectorization}

This section describes our algorithm for  raster plan vectorization.
It consists of three main parts --- preprocessing, vectorization, and postprocessing.
Each part performs its own function to obtain the final vector image.

\subsubsection{Preprocessing}

The main idea of this part is to prepare the image for subsequent vectorization.
Preprocessing is necessary to ensure that the final image has no unnecessary parts, such as interior items.

The algorithm of preprocessing converts each pixel of a black and white raster RGB image to monochrome.
Then it filters pixels by a threshold value.
In the final part of the preprocessing algorithm, the reverse conversion from monochrome to RGB occurs.
As result of the preprocessing algorithm, the image becomes a clean black and white plan.

\subsubsection{Vectorization Algorithm}

Vectorization algorithm converts the black and white raster preprocessed plan to a vector image using the Shi-Tomasi corner detection method~\cite{shi1994good}.

First, there is search for corners in the image.
Shi-Tomasi corner detection method searches for angles based on the colour difference.
This method provides a list of wall corners.
We average the corners coordinates so that they are on the same straight lines, also solving the problem of crooked walls.

After that, a cycle begins, and at each iteration it creates rectangles that can match the walls based on the corners list.
The rectangles are selected based on their similarity to the original image.
Caching and preprocessing of the original bitmap image is implemented for fast filtering.

After suitable rectangles, which correspond to the walls in the initial raster plan, are selected, they are converted into vector paths. 
This algorithm part results in a vector plan corresponding to the initial raster plan.

\subsubsection{Postprocessing}

This part is critical as it allows obtaining the final vector plan with a correct SVG structure and without redundant paths.

The postprocessing algorithm contains two parts.
The first part removes useless paths inside other paths.
The second part combines paths that share a common part and form a rectangle together.
These parts are necessary to preserve the structure of the image with each path responsible for a separate wall of the plan.

The postprocessing algorithm results in a clean structured vector plan.
Each path in the vector plan corresponds to a wall in the original bitmap image.

\section{Comparisons and Experiments}\label{sec:tests}

The experiments include two parts --- comparison with the complex generator and vectorizer approach and LLM approach.
We conducted about $30$ experiments with other prompts and got a similar distribution of metrics.

\begin{table}[H]
\centering
\caption{Comparison of generation with one text prompt of our algorithm with existing ones. The table shows metrics: number of paths, operation time and CLIPScore.}

\begin{tblr}{
  width=0.47\textwidth,
  hline{1-9},
  colspec={|X[2]|X[0.6]|X[1]|X[1.2]|},
}

Model            & Paths & Time, s. & CLIPScore \\
SD + DiffVG      & 512   & 2356.6   & 0.241\\
AuraFlow + LIVE  & 16    & 1971.4   & 0.243\\
SDXL + EvoVec    & 37    & 778.6    & 0.268\\
FLUX + SvgTracer & 94    & 19.2     & 0.261\\
DeepSeek-v3      & 12    & 8.5      & 0.232\\
GPT-4            & 9     & 6.7      & 0.247\\
Our approach     & 12    & 23.3     & 0.288\\
\end{tblr}
\label{tab:comp all}
\end{table}

\subsection{Comparison with Generator and Vectorizator Approach}

We compare our algorithm with the existing approaches based on a generator and a vectorizer.
About $20$ prompts describing apartment plans were generated for the model comparison and $10$ plans were generated based on each prompt to obtain average metrics using models: Stable Diffusion, FLUX, AuraFlow, Stable Diffusion XL and Stable Diffusion XL with white loss.
Then an image is provided to the input of vectorization algorithm.
Compared vectorization algorithms are DiffVG ($N=512$), LIVE ($N=16$), EvoVec, SvgTracer and our algorithm, where $N$ is the number of paths in the final vector image.

To evaluate the visual correspondence between the generated image and the text prompt, we employ the CLIPScore metric~\cite{hessel2021clipscore} and Image Reward metric~\cite{xu2023imagereward}.
However, the Image Reward metric proved to be ineffective in capturing comparative quality, its results are irrelevant and unlikelihood.

Table~\ref{tab:comp all} shows that our algorithm produces final vector images that demonstrate the best correspondence to the prompts they are based on.
Beside, its operation time is one of the shortest compared to others.
In our algorithm, paths number corresponds to the number of walls.

\subsection{Comparison with LLM Approach}

The following is a comparison of our method with such LLMs-based approaches as GPT-4~\cite{achiam2023gpt} and DeepSeek-v3~\cite{liu2024deepseek}.
The main parameters for comparison are the number of paths, operation time and correspondence to the text prompt based on CLIPScore and visual representation.

Table~\ref{tab:comp all} shows a low visual quality of vector plan generation by LLMs.
Although the operation time of LLMs is short and the paths number is small, low compliance with prompt makes such generation inapplicable for our task at the moment.

\section{Conclusion}\label{sec:concl}

In this paper we have presented the novel method for generating vector residential plans based on textual description. 
We designed a new loss function to generate raster images on a white background.
Additionally, we have developed a new vectorization algorithm based on the Shi-Tomasi corner detection method.
Table~\ref{tab:comp all} shows that our algorithm works fast, generates better quality plans compared with all existing approaches with clean SVG structure without unnecessary paths.

\section{Acknowledgments}

This work supported by the Ministry of Economic Development of the Russian Federation (IGK 000000C313925P4C0002), agreement No139-15-2025-010

\bibliography{aaai2026}


\end{document}